\documentclass[sigconf,nonacm]{acmart}
\usepackage{microtype}
\usepackage{pgfplots}
\usepgfplotslibrary{groupplots}
\usepackage{subcaption}
\usepackage{xcolor}
\usepackage{enumitem}

\definecolor{darkred}{HTML}{B24020}
\definecolor{amber}{HTML}{E7A33E}
\definecolor{green}{HTML}{44712E}

\AtBeginDocument{%
  }

\begin{document}

\title{CADET: Context-Conditioned Ads CTR Prediction With a Decoder-Only Transformer}


\author{David Pardoe}
\authornote{Authors contributed equally to this research.}
\author{Neil Daftary}
\authornotemark[1]
\author{Miro Furtado}
\authornotemark[1]
\author{Aditya Aiyer}
\authornotemark[1]
\affiliation{%
  \institution{LinkedIn}
  \city{Mountain View}
  \state{California}
  \country{USA}
}
\email{dpardoe@linkedin.com}

\author{Yu Wang}
\author{Liuqing Li}
\author{Tao Song}
\author{Lars Hertel}
\authornote{Work done while at LinkedIn.}
\affiliation{%
  \institution{LinkedIn}
  \city{Mountain View}
  \state{California}
  \country{USA}
}
\email{ywang11@linkedin.com}

\author{Young Jin Yun}
\author{Senthil Radhakrishnan}
\author{Zhiwei Wang}
\authornotemark[2]
\author{Tommy Li}
\affiliation{%
  \institution{LinkedIn}
  \city{Mountain View}
  \state{California}
  \country{USA}
}
\email{yyun@linkedin.com}

\author{Khai Tran}
\author{Ananth Nagarajan}
\author{Ali Naqvi}
\author{Yue Zhang}
\affiliation{%
  \institution{LinkedIn}
  \city{Mountain View}
  \state{California}
  \country{USA}
}
\email{khtran@linkedin.com}

\author{Renpeng Fang}
\author{Avi Romascanu}
\author{Arjun Kulothungun}
\authornotemark[2]
\author{Deepak Kumar}
\affiliation{%
  \institution{LinkedIn}
  \city{Mountain View}
  \state{California}
  \country{USA}
}
\email{refang@linkedin.com}

\author{Praneeth Boda}
\author{Fedor Borisyuk}
\author{Ruoyan Wang}
\authornotemark[1]
\authornote{Corresponding author.}
\affiliation{%
  \institution{LinkedIn}
  \city{Mountain View}
  \state{California}
  \country{USA}
}
\email{rnwang@linkedin.com}

\renewcommand{\shortauthors}{Pardoe and Daftary, et al.}

\begin{abstract}
Click-through rate (CTR) prediction is fundamental to online advertising systems. While Deep Learning Recommendation Models (DLRMs) with explicit feature interactions have long dominated this domain, recent advances in generative recommenders have shown promising results in content recommendation. However, adapting these transformer-based architectures to ads CTR prediction still presents unique challenges, including handling post-scoring contextual signals, maintaining offline-online consistency, and scaling to industrial workloads.

We present CADET (Context-Conditioned Ads Decoder-Only Transformer), an end-to-end decoder-only transformer for ads CTR prediction deployed at LinkedIn. Our approach introduces several key innovations: (1) a context-conditioned decoding architecture with multi-tower prediction heads that explicitly model post-scoring signals such as ad position, resolving the chicken-and-egg problem between predicted CTR and ranking; (2) a self-gated attention mechanism that stabilizes training by adaptively regulating information flow at both representation and interaction levels; (3) a timestamp-based variant of Rotary Position Embedding (RoPE) that captures temporal relationships across timescales from seconds to months; (4) session masking strategies that prevent the model from learning dependencies on unavailable in-session events, addressing train-serve skew; and (5) production engineering techniques including tensor packing and custom FlashAttention kernels that enable efficient training and serving at scale.

In online A/B testing, CADET achieves an 11.04\% CTR lift compared to the production LiRank~\cite{Borisyuk2024LiRank} baseline model, a hybrid ensemble of DCNv2~\cite{Wang2021DCNv2} and sequential encoders. The system has been successfully deployed on LinkedIn's advertising platform, serving the main traffic for homefeed sponsored updates.
\end{abstract}

\ccsdesc[500]{Information systems~Computational advertising}
\ccsdesc[300]{Computing methodologies~Neural networks}
\ccsdesc[300]{Computing methodologies~Deep learning}
\ccsdesc[100]{Information systems~Online advertising}

\keywords{Click-Through Rate Prediction, Generative Recommenders, Online Advertising}


\maketitle

\section{Introduction}

LinkedIn is the world's largest professional social network, serving over one billion members globally, with a leading B2B-focused advertising platform. Homefeed sponsored updates represent the primary surface for LinkedIn ads, where click-through rate (CTR) prediction serves as the core model powering ad delivery and determining which ads are displayed to users. The accuracy and efficiency of CTR prediction therefore directly impact both LinkedIn user experience and advertiser outcomes.

Historically, Deep Learning Recommendation Models (DLRMs) such as DCN~\cite{Wang2017DCN, Wang2021DCNv2}, DIN~\cite{Zhou2018DIN}, and DHEN~\cite{Zhang2022DHEN}, which focus on explicit feature interaction learning, have dominated ads CTR prediction. These models typically operate on individual ad impressions with a large number of engineered features. Transformer-based sequence encoders have been integrated within DLRM architectures and successfully deployed in industrial systems~\cite{Xia2023TransAct, Borisyuk2024LiRank, Zeng2025InterFormer}. More recently, HSTU-style generative recommenders~\cite{Zhai2024HSTU} reformulated recommendation as a generative paradigm with decoder-only transformers as the sole backbone, and subsequent work such as OneRec~\cite{Deng2025OneRec} and MTGR~\cite{Han2025MTGR} extends this paradigm to industrial-scale retrieval and ranking. A related challenge across ranking systems is handling post-scoring contextual signals such as ad position, which are unavailable at scoring time but influence user behavior; existing approaches treat position as a debiasing feature, apply counterfactual learning~\cite{Agarwal2019Counterfactual}, or employ secondary factorized models~\cite{Anil2022Factory}.

However, adapting decoder-only transformers to production-grade ads CTR prediction still presents unique challenges. First, post-scoring contextual signals such as ad position significantly influence CTR but are unavailable at scoring time, creating a chicken-and-egg problem between predicted CTR and final ranking. Second, online recommendation systems require strict offline-online consistency, yet tracking delays and session-level scoring create gaps between training and serving conditions. Third, production environments demand training stability and reproducibility. Fourth, industrial workloads require efficient training and inference over billions of impressions daily.

In this paper, we present CADET (Context-Conditioned Ads Decoder-Only Transformer), an end-to-end decoder-only transformer for ads CTR prediction deployed at LinkedIn. Our approach addresses the above challenges through novel architectural innovations and production engineering techniques.

Our key contributions are as follows:
\begin{itemize}[noitemsep,topsep=0pt]
\item \textbf{Context-conditioned decoding architecture}: A novel multi-tower decoding block that explicitly models post-scoring contextual signals (e.g., ad position) through $K$ prediction heads, resolving the ranking dependency loop without iterative re-scoring.
\item \textbf{Self-gated attention mechanism}: An improved self-attention design with representation-level and interaction-level gating that stabilizes training and mitigates pathological attention behaviors, enabling reliable convergence at scale.
\item \textbf{Temporal encoding with modified RoPE}: A timestamp-based rotary positional encoding that captures time differences ranging from seconds to months, replacing sequence position with event timestamps for ads-specific temporal modeling.
\item \textbf{Session masking for offline-online consistency}: Customized masking strategies that prevent the model from attending to unavailable in-session events during training, addressing train-serve skew caused by tracking delays.
\item \textbf{Production engineering at scale}: A comprehensive set of optimizations including tensor packing and custom FlashAttention kernels for multi-item scoring that enable efficient training on large-scale datasets and low-latency serving.
\end{itemize}

\begin{figure*}[t]
  \centering
  \begin{subfigure}[t]{0.48\textwidth}
    \centering
    \includegraphics[width=\textwidth]{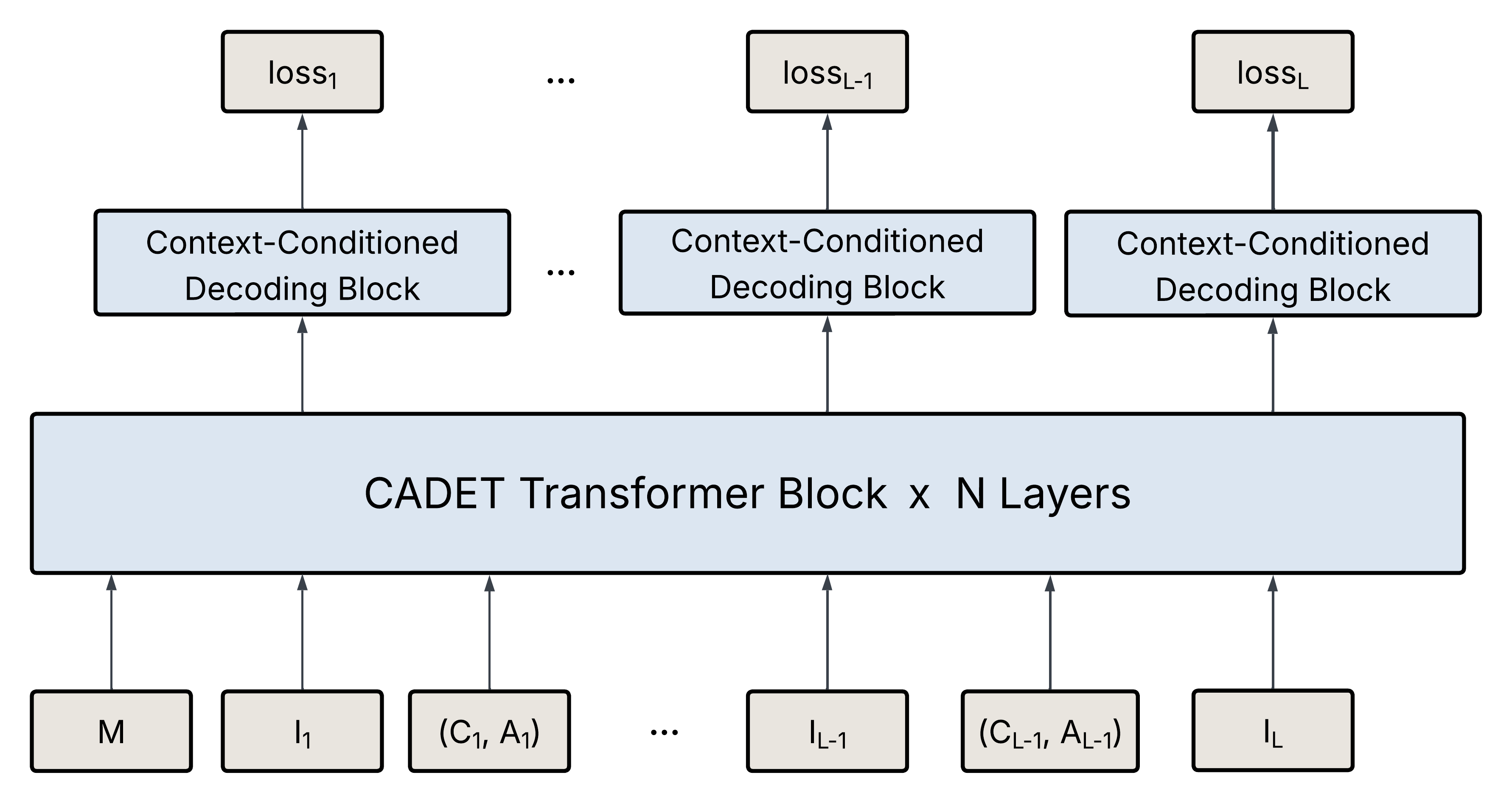}
    \caption{Overall architecture}
    \label{fig:architecture_main}
  \end{subfigure}
  \hfill
  \begin{subfigure}[t]{0.48\textwidth}
    \centering
    \includegraphics[width=\textwidth]{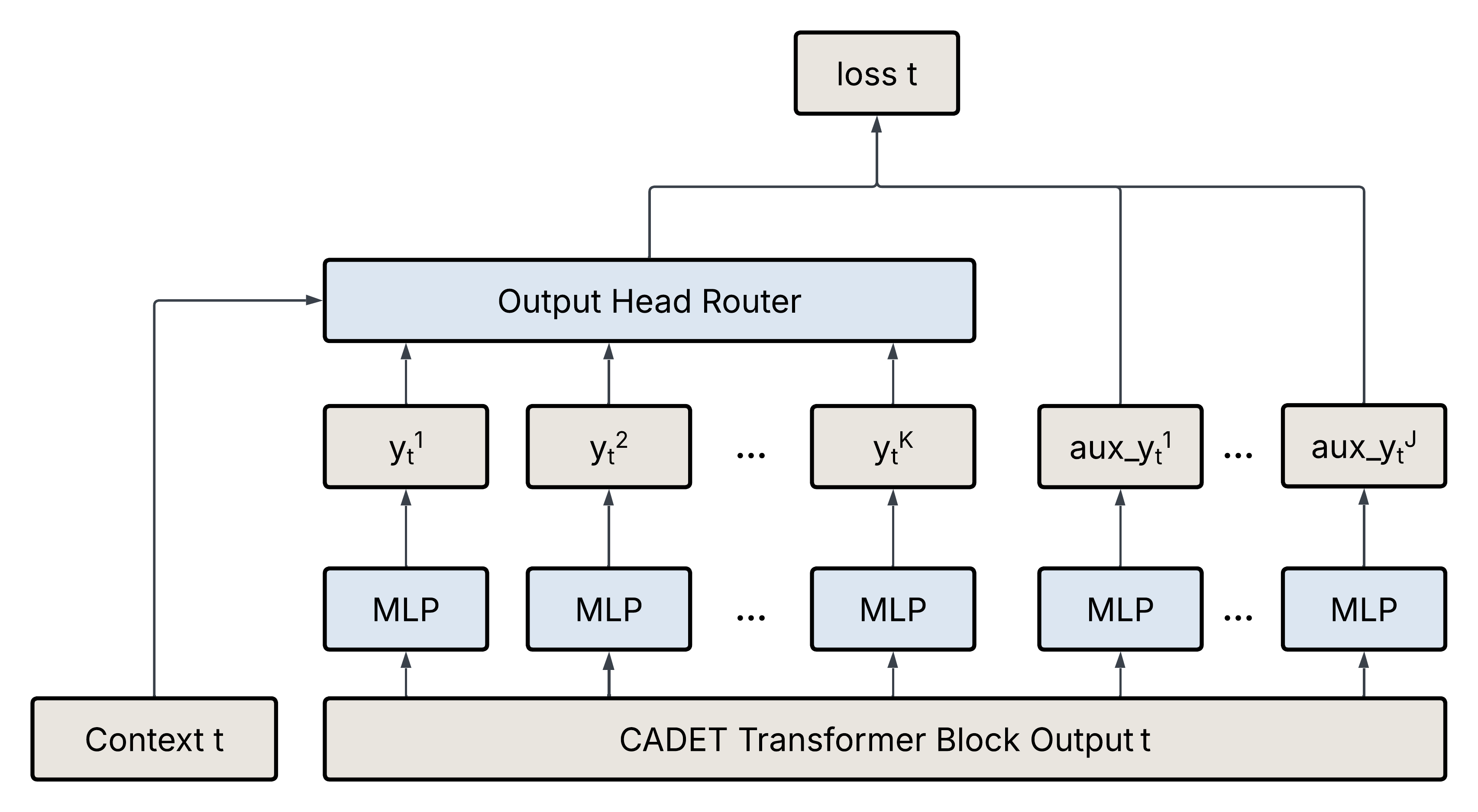}
    \caption{Context-conditioned decoding block}
    \label{fig:architecture_decoding}
  \end{subfigure}
  \caption{Architecture of the proposed model. (a) Overall architecture showing the interleaved impression-action sequence processing through the decoder-only transformer. (b) Detailed view of the context-conditioned decoding block with multiple prediction heads.}
  \label{fig:architecture}
\end{figure*}

\section{Enhancing Decoder-Only Transformer for Ads System}

\subsection{Overall Architecture}
\label{sec:architecture}

Inspired by the HSTU ranking architecture~\cite{Zhai2024HSTU}, the full model processes interleaved ad impression and action sequences:

\paragraph{Input Sequence Formulation:}
The input consists of static user features followed by interleaved impression and contextualized action tokens:
\begin{align}
\mathbf{x} = [M; I_1, (C_1, A_1); I_2, (C_2, A_2); \ldots; I_{L-1}, (C_{L-1}, A_{L-1}); I_L]
\end{align}
where $M$ is user static features (e.g., profile), $I_t$ concatenates ad, request, and other impression features, $C_t$ is post-scoring contextual features unavailable at scoring time (e.g., ad position), and $A_t$ is the action taken on impression $t$ (e.g., click). A decoder-only transformer processes this sequence and predicts the action for every impression in a single pass via teacher forcing.

\paragraph{Context-Conditioned Decoding Block:}
The decoder-only transformer outputs a sequence of predictions paired with contextual metadata for each impression. For each impression, the decoding block produces $K$ context-conditioned predictions:
\begin{align}
\{\hat{y}_1, \hat{y}_2, \ldots, \hat{y}_K\}
\end{align}
where $\hat{y}_k$ represents the prediction for context bucket $k$ (e.g., different ad positions). Through this context-conditioned decoding block, described in Section~2.3, the model learns to produce predictions conditioned on post-scoring contextual information.

\subsection{CADET Transformer Block}

\subsubsection{Self-Gated Multi-Head Attention}

\begin{figure}[t]
  \centering
  \includegraphics[width=0.48\textwidth]{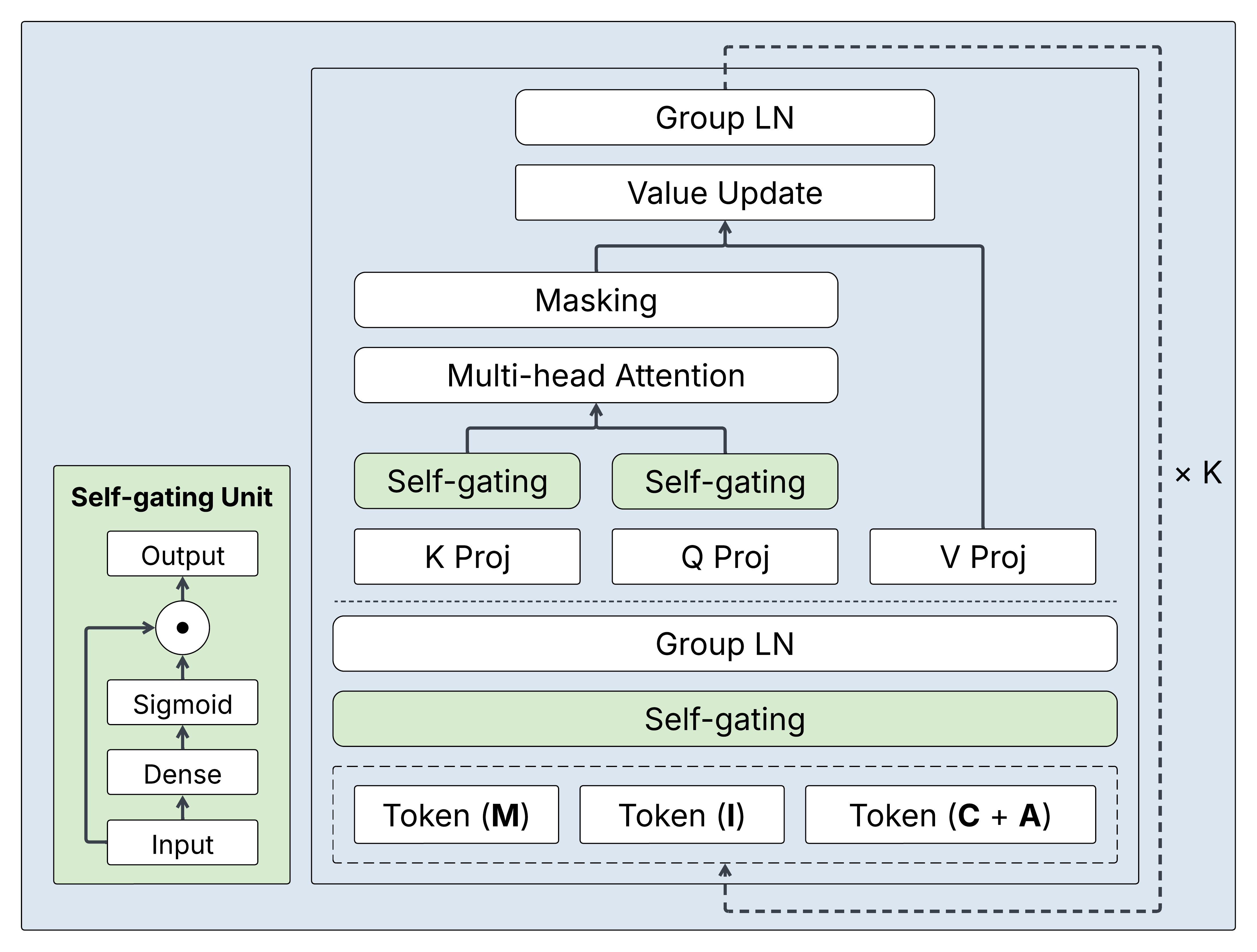}
  \caption{Detailed description of self-gated attention module}
  \label{fig:architecture_gating}
\end{figure}

To improve training stability and mitigate pathological attention behaviors in our proposed model, we introduce self-gated attention \cite{chai-etal-2020-highway} that adaptively regulates information flow at both representation and interaction levels; see Figure \ref{fig:architecture_gating}. Unlike output-level gating used in prior work~\cite{Zhai2024HSTU, Qiu2025GatedAttention}, our approach applies gating \emph{before} attention computation on input representations and query-key projections.

Specifically, let $X \in \mathbb{R}^{B \times L \times d_{\text{model}}}$ denote the token representations at each layer of the decoder-only transformer, where $B$ is batch size, $L$ is sequence length and $d_{\text{model}}$ is the model dimension. Queries $Q$ and keys $K$ are obtained through linear projections of $X$ and preserve the same dimensionality, i.e., $Q, K \in \mathbb{R}^{B \times L \times d_{\text{model}}}$
\begin{equation}
Q = XW_Q \quad K = XW_K,
\end{equation}
where $W_Q, W_K \in \mathbb{R}^{d_{\text{model}} \times d_{\text{model}}}$.
We first apply a lightweight self-gating module to obtain gated representations, where the gate is conditioned on the representations themselves. This \textbf{representation-level gating} acts as adaptive feature selection, suppressing noisy or low-utility dimensions before attention computation, thereby improving activation scaling and gradient conditioning during training.
\begin{equation}
\begin{aligned}
Gate(X) = \sigma(W_X^{\text{gate}}X), \quad 
\tilde{X} = X \odot Gate(X)
\end{aligned}
\end{equation}
where $\sigma(\cdot)$ is the sigmoid function, $\odot$ denotes element-wise multiplication, and $W_X^{\text{gate}} \in \mathbb{R}^{d_{\text{model}} \times d_{\text{model}}}$.

In addition, we incorporate self-gating directly into the attention mechanism by modulating the projected queries $Q$ and keys $K$. 
This \textbf{interaction-level gating} explicitly regulates query-key interactions by constraining dot-product magnitudes, preventing dominant tokens from monopolizing attention and effectively mitigating attention sink phenomena.
\begin{equation}
\begin{aligned}
Gate(Q) = \sigma(W_Q^{\text{gate}}Q), \quad 
\tilde{Q} = Q \odot Gate(Q) \\ 
Gate(K) = \sigma(W_K^{\text{gate}}K), \quad 
\tilde{K} = K \odot Gate(K)
\end{aligned}
\end{equation}
where $W_Q^{\text{gate}}, W_K^{\text{gate}} \in \mathbb{R}^{d_{\text{model}} \times d_{\text{model}}}$.

\subsubsection{Temporal Encoding}

We encode temporal information using a modified form of Rotary Positional Embeddings (RoPE)~\cite{Rope}. In the original RoPE formulation, a query or key vector $\mathbf{x} \in \mathbb{R}^{d}$ is decomposed into $d/2$ two-dimensional components, and the $i$-th component is rotated by an angle
\[
\alpha_i(p) = p \cdot \theta_i,
\qquad
\theta_i = 10000^{-\frac{2i}{d}},
\]
where $p$ denotes the item’s sequence position. The dot product of rotated queries and keys then depends on their angular difference, letting the model learn relative position.

For ads modeling, we care more about differences in time than sequence position, so we replace position $p$ with a Unix timestamp $t$. To handle time differences ranging from seconds to months, we define the rotation angle for the $i$-th component as
\[
\alpha_i(t) = t \cdot \theta_i,
\qquad
\theta_i = \frac{\phi_{min}}{\Delta t_{\max}} \cdot base^{\frac{2i}{d}},
\]
where $\Delta t_{\max}$ is the lookback window used in constructing sequences and thus the largest possible difference in timestamps. $\phi_{min}$ is a tunable parameter that lets us control what the smallest relative rotation (difference in $\alpha_0$) would be for items with timestamps differing by $\Delta t_{\max}$, and $base$ is a tunable parameter that lets us control the highest frequency rotations that are useful for representing short-term information.

We do not add explicit information about sequence position, as causal transformers can learn this implicitly~\cite{Nope}. Crucially, this does not discard ordering: the causal mask still enforces sequence order, so within a dense burst of near-simultaneous events the model retains order while RoPE correctly contributes near-zero relative rotation.

\subsubsection{Customized Masking}

We employ customized time-based masking strategies during training and inference to ensure offline-online consistency and efficient candidate scoring. Here, ``in-session'' denotes not a hard session boundary but events within an impression's delay window $\Delta_{\text{delay}}$ ($t_i-\Delta_{\text{delay}}<t_j\le t_i$), which may not yet be observable at serving time; the masking below depends only on this threshold.

\paragraph{Inference-time Masking.}
At scoring time, the ranking model needs to rank hundreds of candidates simultaneously. Similar to previous work on generative recommenders~\cite{Zhai2024HSTU}, instead of scoring candidates individually, we append all candidates to the end of the sequence and employ a special mask as shown in Figure~\ref{fig:session_masks} so that all candidates are scored in a single forward pass while remaining causally isolated from each other. This significantly improves evaluation and online inference speed.

\paragraph{Training Masking.}
Maintaining offline-online consistency is a critical challenge. In online ranking, recent activity may be unavailable due to system delay, simultaneous ranking across ad slots, or delayed interactions (e.g., a user clicking an ad seen earlier in the session). Attending to such same-session tokens would inflate offline metrics and create reliance on information unavailable at inference time.

To prevent the model from learning to depend on interactions that are unavailable at inference-time, we enforce session-aware masking during training. We apply a train-time mask $M$ that prevents tokens from attending to same-session information:
\begin{align}
M_{i,j} &=
\begin{cases}
-\infty & \text{if } t_j > t_i - \Delta_{\text{delay}} \\
0 & \text{otherwise}
\end{cases}
\end{align}
where $\Delta_{\text{delay}}$ encodes the minimum latency we guarantee between an event and its observability online. We materialize this mask at runtime as shown in Figure~\ref{fig:session_masks}. The final attention computation is:

\begin{equation}
\text{Attention}(Q,K,V) = \text{softmax}\!\left(\frac{QK^\top}{\sqrt{d_k}} + M\right)V
\end{equation}
\usetikzlibrary{patterns}

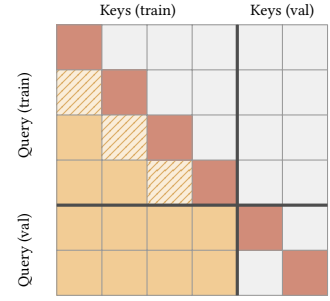
\begin{figure}[t]
  \centering
  \begin{tikzpicture}[x=0.6cm,y=0.6cm]
    \def\Train{4}
    \def\Total{6}
    \def\Delay{2}
    \pgfmathtruncatemacro{\Last}{\Total-1}
    \pgfmathtruncatemacro{\SplitY}{\Total-\Train} 

    \foreach \row in {0,...,\Last} {
      \foreach \col in {0,...,\Last} {

        \pgfmathtruncatemacro{\trainallowed}{
          (
            (\row < \Train) &&
            (
              ( (\row >= \Delay) && (\col <= \row - \Delay) )
              || (\col == \row)
            )
          )
          ||
          (
            (\row >= \Train) &&
            (
              (\col < \Train) || (\col == \row)
            )
          )
        }

        \pgfmathtruncatemacro{\causalallowed}{
          (\row < \Train) && (\col < \Train) && (\col <= \row)
        }

        \pgfmathtruncatemacro{\inferonly}{
          (\causalallowed == 1) && (\trainallowed == 0)
        }

        \ifnum\trainallowed=1
          \fill[amber!55] (\col,{\Last-\row}) rectangle ++(1,1);
        \else
          \ifnum\inferonly=1
            \fill[amber!35,opacity=0.55] (\col,{\Last-\row}) rectangle ++(1,1);
            \fill[
              pattern=north east lines,
              pattern color=amber!90!black,
              opacity=0.85
            ] (\col,{\Last-\row}) rectangle ++(1,1);
          \else
            \fill[gray!12] (\col,{\Last-\row}) rectangle ++(1,1);
          \fi
        \fi

      }
    }

    \foreach \row in {0,...,\Last} {
      \fill[darkred!60] (\row,{\Last-\row}) rectangle ++(1,1);
    }

    \draw[step=1,black!40,thin] (0,0) grid (\Total,\Total);
    \draw[very thick,black!70] (\Train,0) -- (\Train,\Total);

    \draw[very thick,black!70] (0,\SplitY) -- (\Total,\SplitY);

    \node at (1.8,{\Total+0.3}) {\scriptsize Keys (train)};
    \node at (5,{\Total+0.3}) {\scriptsize Keys (val)};

    \node[rotate=90] at (-0.7,4) {\scriptsize Query (train)};
    \node[rotate=90] at (-0.7,1) {\scriptsize Query (val)};
  \end{tikzpicture}
  \caption{Session-aware masking strategies (depicted before item/action interleaving). Solid yellow cells are always unmasked. Striped cells fall within the $\Delta_{\text{delay}}$ threshold, masked during training, but available at inference. Gray cells are masked and dark red cells denote the preserved diagonal.}
  \label{fig:session_masks}
\end{figure}

\subsection{Context-Conditioned Decoding Block}

The decoding block addresses a fundamental challenge in ads recommendation: the gap between training-time observability and inference-time constraints. Critical contextual signals, such as final ad position or specific UI treatments, are often unknown during initial scoring. We resolve this by employing a multi-tower architecture that generates predictions conditioned on potential rendering contexts in a single forward pass.

\subsubsection{Context-Conditioned Heads}

To model behavior influenced by post-scoring context, we partition the context space into $K$ discrete buckets based on historical click-through rates (e.g., $k = 1$ for positions 1--4 and $k = 2$ for positions 5+, the configuration we deploy). For each bucket $k \in \{1, \ldots, K\}$, we define a dedicated prediction head branching from the shared transformer output $h_t$:
\begin{equation}
\hat{y}_k = \text{MLP}_k(h_t)
\end{equation}
where $h_t$ is the transformer representation for impression $t$ and $\hat{y}_k$ denotes the logit produced by the $k$-th head.

\paragraph{Training:} During training, for each impression $t$, we observe its ground-truth context bucket $k_t \in \{1, \ldots, K\}$. Only the head corresponding to the realized context contributes to the loss:
\begin{equation}
\mathcal{L}_{\text{ctx}} = \sum_{t=1}^{L} \mathrm{CE}\!\left(\hat{y}_{k_t},\, y_t\right)
\end{equation}
where $\hat{y}_{k_t}$ denotes the prediction from the $k_t$-th head, and $y_t$ is the ground-truth label. This selective routing ensures each head learns CTR for its specific context while eliminating train-serve mismatch by using position only for loss routing rather than as an input feature.

\paragraph{Serving:} At inference time, all $K$ heads produce predictions in parallel, yielding $\{\hat{y}_1, \ldots, \hat{y}_K\}$ for each candidate ad. During the online auction, when the serving system renders ads for a specific context, it selects the prediction from the corresponding bucket---e.g., $\hat{y}_1$ when filling positions 1--4, or $\hat{y}_2$ when filling positions 5+. This single-pass approach provides policy-ready signals that resolve the ranking dependency loop without the latency cost of iterative re-ranking or multi-pass feature toggling.

\paragraph{Historical Context Information:} As described in Section~2.1, contextual signals $C_t$ are paired with actions $A_t$ rather than included in impression features $I_t$. This design prevents information leakage during training while enabling the model to learn historical behavior patterns conditioned on context.

\subsubsection{Auxiliary Action Heads}
To improve representation learning, we introduce auxiliary prediction tasks for fine-grained user actions, including click type classification, landing-page dwell time, and impression duration regression. Each auxiliary head $\hat{y}^{\text{aux}}_j = \text{MLP}_j(h_t)$ is a shared prediction module used solely for training-time regularization and does not participate in inference.

\subsubsection{Loss Construction}

The total loss combines the context-conditioned and auxiliary heads:
\begin{equation}
\mathcal{L} = \lambda_{\text{ctx}} \mathcal{L}_{\text{ctx}} + \sum_{j=1}^{J} \lambda_j \mathcal{L}_j^{\text{aux}}
\end{equation}
where $J$ is the number of auxiliary tasks, $\mathcal{L}_j^{\text{aux}}$ denotes the auxiliary loss for task $j$, and $\{\lambda_{\text{ctx}}, \lambda_j\}$ are hyperparameters balancing the terms.
\section{Productionization and Scaling}

\subsection{Training System}

We train on NVIDIA H200 GPUs using PyTorch Hybrid Sharded Data Parallel (HSDP) and FlexAttention~\cite{flexattention} to support customized masking.

The autoregressive paradigm let us densify training data. We migrated from a legacy pointwise format, where each impression was an independent record with over 100 features, to a user-centric sequence format; aggregating impressions and actions into unified sequences and pruning to high-utility signals substantially reduces the data footprint while preserving accuracy (see Section~4.2).

\subsubsection{Packing}

A common approach to training sequence models materializes a dense $(B, L, d_{\text{model}})$ tensor, but since attention is the only operation requiring explicit sequence structure, the majority of activation memory and MLP compute is wasted on padding tokens.

To densify training, we make the packed representation the canonical form throughout training and validation. The dataloader compacts every batch into a contiguous token buffer of shape $(\texttt{total\_tokens}, d_{\text{model}})$ together with prefix-sum offsets and sequence-length metadata. Each worker aggregates user sequences until the packed length approaches a configured budget, pads the final few tokens to hit the exact target, and emits the batch---fixing \texttt{total\_tokens} per batch is critical for stability and \texttt{torch.compile}.

\subsection{Serving System}

Each inference request scores multiple candidate ads against a shared user context (Section~\ref{sec:architecture}) in a single forward pass---the multi-item scoring pattern described in Section~2.2.3---within a p99 latency budget of 50\,ms. Because this pattern is simpler than the session masking used during training, we develop a custom FlashAttention~\cite{dao2024flashattention3} kernel that exploits its sparsity, integrating the masking logic directly into the tiled attention computation and avoiding mask materialization. The shared-context attention shape is $L^2/2 + LN$ versus a standard $(L+N)^2$---linear rather than quadratic in the number of candidates.

\section{Experiments and Results}
Here we present the results of offline experiments designed to illustrate the contributions of various components of the CADET model, as well as online A/B test results. All experiments are conducted on LinkedIn's sponsored updates ad click dataset. 

\subsection{Ablation Studies on CADET Components}

Table \ref{tab:ablation_table} shows the results of ablation studies on various model components, measured in relative ROC AUC change compared to the full CADET model. All experiments are performed on one year of data with the final day used for validation and a $\Delta_{\text{delay}}$ of one hour. We use $n_{\text{layers}}=8$, $n_{\text{heads}}=4$, $d_{\text{model}}=352$, and sequence length $L=4096$.

\subsubsection{Self-Gated Multi-Head Attention}

We study the effect of self-gating on training dynamics by comparing a baseline without gating, representation-level gating only, and the combined representation- and interaction-level configuration. Across multiple runs, the baseline exhibits clear instability with sharp early-training AUC drops; representation-level gating smooths the trajectory by reducing the frequency and magnitude of these drops; and the combined configuration yields the most stable convergence.

\subsubsection{Context-Conditioned Modeling}
We use LinkedIn feed position as the context signal in experiments. We set $K=2$ (positions 1--4, and 5+) to simplify integration with the downstream auction system while capturing the primary position-dependent CTR variation. Removing context-conditioned modeling significantly hurts performance, reducing AUC by 0.61\%.

\subsubsection{Temporal RoPE}
We used $\Delta t_{\max} =$ 1 year in milliseconds, $\phi_{min}=10^{-4}$, and $base=6\times10^{5}$. Removing RoPE applied to impression timestamps reduces AUC by 0.13\%, showing that the model is able to learn the importance of time between impressions.

\subsubsection{Session Masking}
In this experiment, we retain the $\Delta_{\text{delay}}$ of one hour when creating attention masks for validation tokens, but set $\Delta_{\text{delay}}$ to zero otherwise, meaning that during training we allow tokens to attend to any previous token. The drop of 0.31\% in AUC shows that the model is indeed harmed by allowing it access to information at training time that would not be available during inference due to latency in updating sequences.

\begin{table}[htbp]
\centering
\small
\begin{tabular}{l|c}
\hline
\textbf{Model Variant} & \textbf{$\Delta$AUC} \\
\hline
full CADET & --- \\
\quad - context-conditioned & $-0.61\%$ \\
\quad - temporal RoPE & $-0.13\%$ \\
\quad - session masking & $-0.31\%$ \\
\hline
\end{tabular}
\caption{Ablation studies on CADET components.}
\label{tab:ablation_table}
\end{table}

We additionally compare against HSTU-style baselines (Table~\ref{tab:hstu_comparison}). \textit{CADET-HSTU} replaces the Transformer backbone with HSTU while keeping the rest of CADET's design (context-conditioned scoring, self-gating, timestamp RoPE); \textit{CADET-Transformer} is our default backbone. Both variants are reported without session masking because session masking is not readily supported by Meta's released HSTU Triton kernel. The two backbones perform within $0.0001$ AUC of each other, indicating CADET's gains come from its modeling components rather than the backbone choice; full CADET (with session masking) reaches $0.8554$.

\begin{table}[htbp]
\centering
\small
\begin{tabular}{l|c}
\hline
\textbf{Model Variant} & \textbf{AUC} \\
\hline
HSTU-Base & 0.8497 \\
HSTU-Large & 0.8505 \\
CADET-HSTU (w/o session masking) & 0.8535 \\
CADET-Transformer (w/o session masking) & 0.8534 \\
Full CADET & \textbf{0.8554} \\
\hline
\end{tabular}
\caption{Comparison with HSTU baselines and backbone-swapped CADET variants.}
\label{tab:hstu_comparison}
\end{table}

\subsection{Scaling and Efficiency}

Migrating from the legacy pointwise format to user-centric sequences with 12 high-utility signals reduced our data footprint by 98\% while preserving accuracy through prioritization of temporal behavioral signals. Packing eliminates padding tokens, increasing effective batch size by approximately 4$\times$ given the right-skewed user-sequence distribution. Our custom FlashAttention kernel achieves a 3$\times$ speedup over fused multi-head attention (FMHA), reaching 330 requests per second per GPU at p99~$<$~50\,ms.

\subsection{Online Results}

We evaluated CADET through large-scale online A/B testing on LinkedIn's homefeed sponsored updates against the production \textbf{LiRank}~\cite{Borisyuk2024LiRank} baseline---a hybrid ensemble of \textbf{DCNv2}~\cite{Wang2021DCNv2} for feature interactions and \textbf{TransAct}~\cite{Xia2023TransAct} for sequential history modeling, which we replaced entirely with the unified CADET model. CADET achieved a statistically significant \textbf{+11.04\%} lift in CTR over the LiRank production baseline, demonstrating that a unified generative approach can outperform a highly optimized multi-component hybrid system. Because LinkedIn advertisers typically spend their full budgets under auto-bidding, revenue remained approximately neutral. These relevance gains translated into stronger advertiser ROI, with cost-per-click (CPC) falling 10.9\%.

\section{Conclusion}
We presented CADET, an end-to-end decoder-only transformer for large-scale ads CTR prediction at LinkedIn. This work addressed key challenges in applying autoregressive transformers to advertising: self-gated attention stabilizes optimization, timestamp-based RoPE captures temporal relationships across timescales, session-aware masking ensures offline-online consistency, and context-conditioned decoding resolves the chicken-and-egg problem between CTR prediction and post-scoring signals.

Beyond modeling innovations, productionization techniques including tensor packing and custom FlashAttention kernels enable efficient training and low-latency serving. Empirically, CADET achieves 11.04\% CTR lift over our LiRank baseline~\cite{Borisyuk2024LiRank}, showing that a unified decoder-only transformer, combined with targeted modeling and production optimizations, can effectively outperform DLRM-style ensemble models in industrial ads ranking.

We leave to future work the extension to multiple concurrent post-scoring signals beyond position and evaluation on public benchmarks that include such context.

\begin{acks}
We thank Oren Sar Shalom, Yuan Gao and Franklin Graves for their valuable discussions and paper reviews. We also thank Guandong Zhu, Rutvij Mehta, Serhat Leloglu, Zeyu Jin, Shuzhe Xiao, Oz Ozkan, Hina Arora, Shao Tang, Alex Berlingeri, Ankur Agrawal, Emily Ki, Hongzhou Li, Joe Cho, Qian Li, Siva Popuri, Tiffany Zhou, Chen Zhu, Haoyue Tang, Jaideep Ray, Yang Pei, Yujie Ai, Sumit Rangwala, and Xiaoling Zhai for their collaboration and contributions to this work. Finally, we thank Sen Zhou, Xianxing Zhang, Julie Choi, Manas Apte, Yitong Zhou, and Sriram Sankar for their guidance and leadership support.
\end{acks}

\bibliographystyle{ACM-Reference-Format}
\bibliography{references}

\appendix
\section*{APPENDIX}
\section{Experimental Setup}
\label{app:setup}

This appendix consolidates the model, data, training, and evaluation configuration used throughout the paper.

\paragraph{Data.} All experiments use LinkedIn's homefeed sponsored-updates ad-click dataset. Unless otherwise noted, we train on approximately one year of interactions and hold out the final day for validation, using a delay threshold $\Delta_{\text{delay}}$ of one hour. Each user is encoded as static features $M$ followed by an interleaved sequence of impression and contextualized-action tokens; we retain 12 high-utility signals per token, down from over 100 features in the legacy pointwise format.

\paragraph{Model.} CADET is a decoder-only transformer with $n_{\text{layers}}=8$, $n_{\text{heads}}=4$, model dimension $d_{\text{model}}=352$, and maximum sequence length $L=4096$. Every layer applies self-gated attention with both representation- and interaction-level gating. Temporal information uses the timestamp-based RoPE with $\Delta t_{\max}=1$~year (in milliseconds), $\phi_{min}=10^{-4}$, and $base=6\times10^{5}$. The context-conditioned decoding block uses $K=2$ position buckets (positions 1--4 and 5+). Auxiliary heads predict click type, landing-page dwell time, and impression duration, and are used only at training time. The training objective is $\mathcal{L}=\lambda_{\text{ctx}}\mathcal{L}_{\text{ctx}}+\sum_{j}\lambda_j\mathcal{L}_j^{\text{aux}}$.

\paragraph{Training.} Models are trained on NVIDIA H200 GPUs with PyTorch Hybrid Sharded Data Parallel (HSDP) and FlexAttention~\cite{flexattention} to support session-aware masking. Each batch uses the packed $(\texttt{total\_tokens}, d_{\text{model}})$ representation with a fixed token budget, which is required for stable \texttt{torch.compile}.

\paragraph{Serving and evaluation.} At inference, each request scores multiple candidate ads against a shared user context in a single forward pass, served by a custom FlashAttention kernel within a p99 latency budget of 50\,ms ($\sim$330 requests per second per GPU). This kernel is purely a serving-latency optimization: all training and offline evaluation use stock FlexAttention, and none of the accuracy results in the paper depend on it, so CADET's modeling contributions reproduce without any bespoke attention code. Offline quality is reported as relative ROC~AUC, and ablations as relative $\Delta$AUC against the full CADET model.

\end{document}